\documentclass[10pt,twocolumn,letterpaper]{article}

\usepackage{cvpr}
\usepackage{times}
\usepackage{epsfig}
\usepackage{graphicx}
\usepackage{amsmath}
\usepackage{amssymb}
\usepackage{textcomp}

\usepackage{subcaption}
\usepackage{multirow}
\usepackage{adjustbox}
\usepackage{tabu}
\usepackage{xcolor}
\usepackage{soul}
\sethlcolor{lightgray}

\usepackage[pagebackref=true,breaklinks=true,letterpaper=true,colorlinks,bookmarks=false]{hyperref}

\cvprfinalcopy 

\def\cvprPaperID{2343} 

\ifcvprfinal\fi
\begin{document}

\title{EANet: Enhancing Alignment for Cross-Domain Person Re-identification}
\author{Houjing Huang$^{\,1}$ \ \ \ Wenjie Yang$^{\,1}$ \ \ \ Xiaotang Chen$^{\,1}$ \ \ \ Xin Zhao$^{\,1}$ \ \ \ Kaiqi Huang$^{\,1}$\\
Jinbin Lin$^{\,2}$ \ \ \ Guan Huang$^{\,2}$ \ \ \ Dalong Du$^{\,2}$\\
$^{1}$ CRISE \& CASIA \ \ \ $^{2}$ Horizon Robotics, Inc.\\
}

\maketitle

\begin{abstract}
   Person re-identification (ReID) has achieved significant improvement under the single-domain setting. However, \textbf{directly exploiting} a model to new domains is always faced with huge performance drop, and \textbf{adapting} the model to new domains without target-domain identity labels is still challenging. In this paper, we address cross-domain ReID and make contributions for both model \textbf{generalization} and \textbf{adaptation}. First, we propose Part Aligned Pooling (PAP) that brings significant improvement for cross-domain testing. Second, we design a Part Segmentation (PS) constraint over ReID feature to enhance alignment and improve model generalization. Finally, we show that applying our PS constraint to unlabeled target domain images serves as effective domain adaptation. We conduct extensive experiments between three large datasets, Market1501, CUHK03 and DukeMTMC-reID. Our model achieves state-of-the-art performance under both source-domain and cross-domain settings. For completeness, we also demonstrate the complementarity of our model to existing domain adaptation methods. The code is available at \textcolor{magenta}{\url{https://github.com/huanghoujing/EANet}}.
\end{abstract}


\section{Introduction}

\begin{figure}
    \centering
    \begin{subfigure}{0.5\linewidth}
        \centering
        \includegraphics[width=0.98\linewidth]{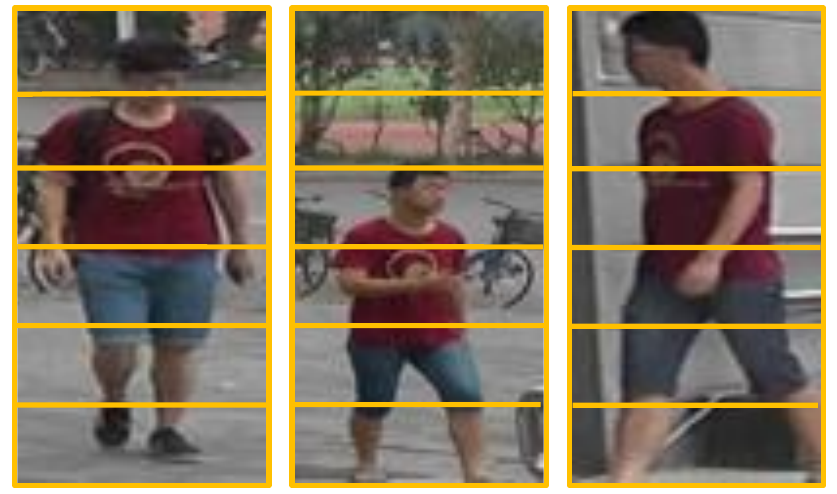}
        \caption{PCB Pooling Regions}
        \label{fig:pcb_pap_regions_a}
    \end{subfigure}%
    \begin{subfigure}{0.5\linewidth}
        \centering
        \includegraphics[width=0.98\linewidth]{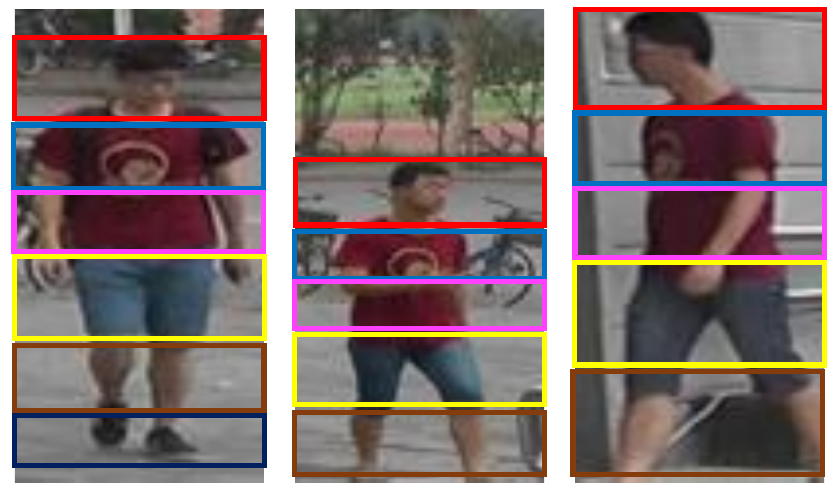}
        \caption{Part Aligned Pooling Regions}
        \label{fig:pcb_pap_regions_b}
    \end{subfigure}%
    \\
    \begin{subfigure}{1\linewidth}
        \centering
        \includegraphics[width=1\linewidth]{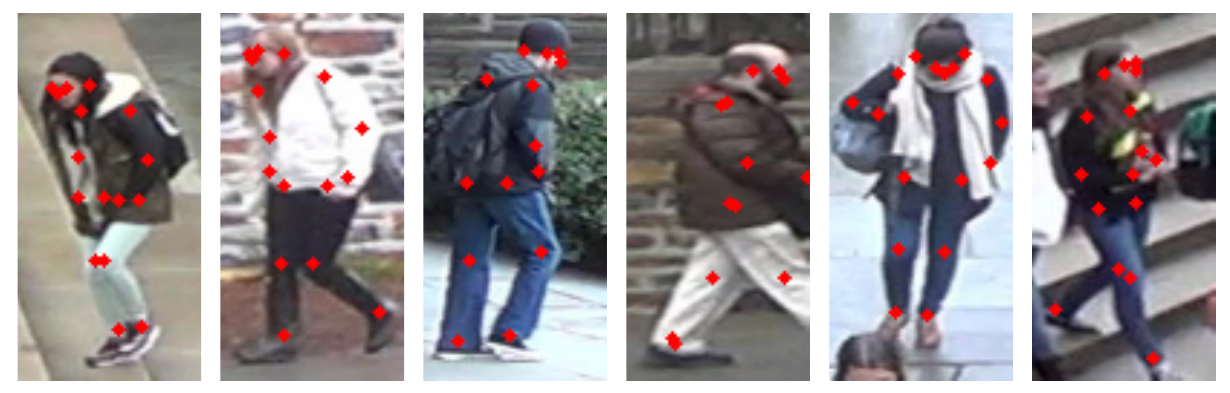}
        \caption{Used Keypoints}
        \label{fig:pcb_pap_regions_c}
    \end{subfigure}
    \caption{(a) PCB~\cite{sun2017beyond} pools feature from evenly divided stripes. (b) We pool feature from keypoint delimited regions (The other three regions are shown in Figure~\ref{fig:model}). (c) The keypoints predicted by a model trained on COCO~\cite{lin2014microsoft}, for determining regions in (b).}
	\label{fig:pcb_pap_regions}
\end{figure}

Person re-identification is a fundamental task in video surveillance that serves pedestrian retrieval and cross-camera tracking~\cite{zheng2016person,ristani2018features}, \etc. It aims to predict whether two images from different cameras belong to the same person. With large scale datasets, as well as improved feature extraction and metric learning methods, recent years have seen great improvement in this task under the single-domain setting.

However, there is a huge performance drop when the model is directly exploited in unseen domains, which is apparent when testing the model on new datasets~\cite{deng2018image,wei2017person,wang2018transferable}. It's partially due to the zero-shot setting of ReID, that test identities are never seen during training. The data distribution discrepancy between domains may also matter, \eg Market1501 contains pedestrians usually wearing shorts, while DukeMTMC-reID shows frequent trousers and coats. It would be expensive to build a training set for each scene due to costly cross-camera person grouping~\cite{li2017learning}. Thus, the problem of \textbf{(1)} improving generalization of a model when training on source domain or \textbf{(2)} adapting the model to new domains without any target-domain identity label, is extremely important for cross-domain ReID. Recently, style transfer~\cite{deng2018image,wei2017person}, attribute recognition~\cite{wang2018transferable} and target-domain label estimation~\cite{liu2017stepwise,lv2018unsupervised,li2018unsupervised} have been shown beneficial for overcoming domain shift.

In this paper, we verify that part alignment plays an important role in ReID model generalization. Besides, we show that training unlabeled target-domain images with part segmentation constraint, while training ReID on source domain, serves as an effective way of domain adaptation. \textbf{Firstly}, we improve previous state-of-the-art part model PCB~\cite{sun2017beyond} with the assistance of pose estimation. PCB evenly partitions feature map into $P$ horizontal stripes for pooling local features, as demonstrated in Figure~\ref{fig:pcb_pap_regions_a}. Obviously, it fails to align body parts when imperfect detection occurs. Our alignment operation is pooling part feature from keypoint delimited regions, Figure~\ref{fig:pcb_pap_regions_b}. The keypoint coordinates are obtained from a pose estimation model trained on COCO~\cite{lin2014microsoft}. This alignment strategy increases cross-domain scores prominently. \textbf{Secondly}, we discover that features pooled from adjacent parts share much similarity (Figure~\ref{fig:part_sim}), which we argue indicates (1) missing localization ability and (2) feature redundancy between different regions. It may be caused by the large field of view (FoV) of Conv5 and the strict ReID constraint on each part. We propose to enhance the localization ability of ReID features, by imposing part segmentation constraint on the feature maps. Specifically, we connect a segmentation module to ReID feature maps, training it with pseudo segmentation labels predicted by a model trained on COCO Densepose data~\cite{guler2018densepose}. The simple structure improves model generalization significantly. \textbf{Finally}, for adapting the model to certain domains, we feed the unlabeled target-domain images to the model, training them with the proposed part segmentation constraint. Experiments show that this target-domain regularizer without identity label is really effective. With the mentioned proposals, our model achieves state-of-the-art performance under both source-domain and cross-domain settings.

Our contribution is threefold. \textbf{(1)} We propose part aligned pooling that improves ReID cross-domain testing prominently. \textbf{(2)} We design a part segmentation constraint over ReID feature to further improve model generalization. \textbf{(3)} We propose to apply our part segmentation constraint on unlabeled target-domain images as effective domain adaptation.

\section{Related Work}

\subsection{ReID and Part Based Models}
Person ReID aims to predict whether two images belong to the same person. The testing protocol takes the form of person retrieval. For each query image, we calculate its feature distance (or similarity) to each gallery image and sort the resulting list. The ranking result illustrates the performance of the model. Thus a ReID model mainly involves feature extraction and similarity metric. In terms of \textbf{similarity metric}, representative works including triplet loss~\cite{Schroff_2015_CVPR,hermans2017defense,wang2018mancs}, quadruplet loss~\cite{chen2017beyond}, re-ranking~\cite{zhong2017re,yu2017divide}, \etc. In terms of \textbf{feature extraction}, recent works have paid intensive attention to part based feature, for its multi-granularity property, part attention or alignment. \textbf{Group 1.} The first group did not require keypoint or segmentation information. Li \etal~\cite{li2017learning} utilized STN to localize body parts and extract local features from image patches. Zhao \etal~\cite{zhao2017deeply} proposed a simple attention module to extract regional features emphasized by the model. Sun \etal~\cite{sun2017beyond} proposed a strong part baseline and refined part pooling. \textbf{Group 2.} With the rapid development of pose estimation~\cite{cao2017realtime,xiao2018simple} and human parsing algorithms~\cite{gong2017look}, more and more ReID researchers resort to the assistance of predicted keypoints or part regions. Su \etal~\cite{su2017pose} cropped, normalized and combined body parts into a new image for network input. Kalayeh \etal~\cite{kalayeh2018human} trained a part segmentation model on human parsing dataset LIP~\cite{gong2017look} to predict 4 body parts and foreground. Local region pooling was then performed on feature maps. Xu \etal~\cite{xu2018attention} shared similar idea, but with regions generated from keypoints. Besides, part visibility is also integrated into the final feature. Sarfraz \etal~\cite{sarfraz2017pose} directly concatenated 14 keypoint confidence maps with the image as network input, letting the model learns alignment in an automatic way. Suh \etal~\cite{suh2018part} proposed a two-stream network, a ReID stream and a pose estimation stream and used bilinear pooling to obtain part-aligned feature.

Previous part based methods are with advantages and shortcomings in different aspects, our model is designed with this consideration. It has following merits. (1) We extract part feature from feature maps, sharing the whole backbone for all parts, which is efficient.  (2) The pooling regions are determined by keypoint coordinates, for ensuring alignment. (3) ReID supervision is imposed on each part to achieve discriminative part features. (4) Part visibility is explicitly handled during training and testing.

\subsection{Cross-Domain ReID}

Due to expensive identity labeling for ReID, in practice, we expect a model trained in one domain can be directly exploited, or easily adapted, to new ones, without the need for target-domain identity labeling. Direct exploitation considers model \textbf{generalization}, while adapting using unlabeled target-domain images involves \textbf{unsupervised domain adaptation}. TJ-AIDL~\cite{wang2018transferable} learned an attribute-semantic and identity-discriminative feature space transferrable to unseen domains, with a novel attribute-ReID multi-task learning framework. Besides, its attribute consistency constraint also utilized unlabeled target-domain images for domain adaptation. SPGAN~\cite{deng2018image} and PTGAN~\cite{wei2017person} both employed CycleGAN, yet with different generator constraints, to transfer source-domain images into target-domain style. Then a usual ReID model was trained on these translated images for a model suitable for the target domain. Another line of works \cite{liu2017stepwise,lv2018unsupervised,li2018unsupervised,song2018unsupervised} estimated pseudo identity labels on target domain for supervised learning, usually using clustering methods.

Our method concentrates on alignment for model generalization and adaptation, which is complementary to existing cross-domain methods. To be specific, in our multi-task framework, we can apply attribute recognition as an assistant task for both source-domain training and target-domain adaptation. For style transfer and label estimation strategies, our part aligned model works as a strong ReID model and the parsing constraint is also applicable.

\section{Method}

\begin{figure*}[t]
\begin{center}
   \includegraphics[width=0.97\textwidth]{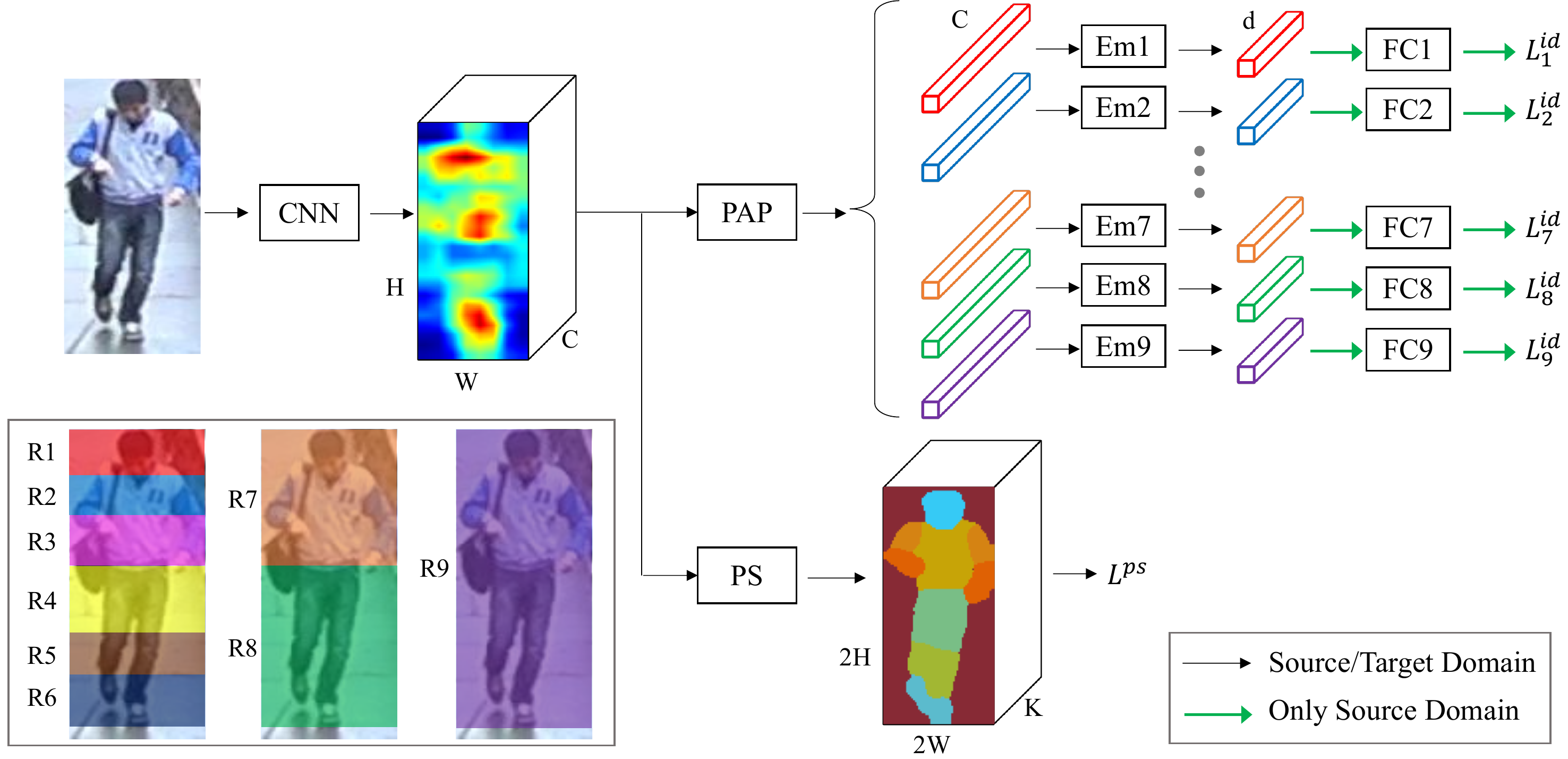}
\end{center}
   \caption{Model Overview. An input image is fed to a CNN to obtain feature maps with shape $C \times H \times W$. Then our Part Aligned Pooling (PAP) module performs max pooling inside each keypoint delimited region. The figure shows 9 regions R1$\sim$R9 on the image. After max pooling, each feature vector is sent to its corresponding embedding layer, one of Em1$\sim$Em9, to reduce dimension from $C$ to $d$. FC1 $\sim$ FC9 and $L_1^\text{id} \sim L_9^\text{id}$ denote identity classifiers and identity classification losses for each part. To enhance alignment, we connect a Part Segmentation (PS) module to the feature maps. It consists of a stride-2 Deconv layer and a $1 \times 1$ Conv layer and is supervised by pixel-wise cross-entropy loss. Since we do not use identity annotation of target domain, we train those images with only segmentation loss.}
\label{fig:model}
\end{figure*}

Our model is depicted in Figure~\ref{fig:model}. It features Part Aligned Pooling (PAP), each-part ReID supervision, and Part Segmentation (PS) constraint, which are detailed in the following.

\subsection{ReID Model with Regional Pooling}

The previous state-of-the-art method PCB demonstrates the effectiveness of pooling local features and imposing identity supervision on each region independently. For an image $I$, we feed it to a CNN and obtain feature maps $G$ with shape $C \times H \times W$, where $C$, $H$ and $W$ are number of channels, height and width, respectively. Suppose there are $P$ spatial regions on $G$ which we extract feature from. From the $p$-th region, we pool a feature vector $g_p \in R^C $, transform it using an embedding layer $e_p = f_p(g_p)$ ($f_p$ means the embedding layer for the $p$-th region), and connect it to the corresponding classifier $W_p$ to classify it into one of the $M$ identities in the training set. The cross-entropy loss $L_p^\text{id}$ is then calculated accordingly. In PCB, loss from different parts are simply accumulated as follows
	\begin{equation}
	L^\text{id}_\text{PCB} = \sum_{p=1}^{P} L_p^\text{id},
	\end{equation}
	and during testing, features from parts are concatenated. Our model also adopts each-part ReID supervision. Besides, we explicitly consider alignment and visibility, as detailed below. 

\subsection{Part Aligned Pooling}

PCB evenly splits feature maps into ${P}$ horizontal stripes and then applies pooling in each stripe to obtain local feature. We argue that this method causes misalignment when imperfect detection occurs. For example, in Figure~\ref{fig:pcb_pap_regions_a}, the same stripes of different images do not correspond to the same body part. 

To achieve part alignment, we propose to pool feature from keypoint delimited regions, as in Figure~\ref{fig:pcb_pap_regions_b} and~\ref{fig:model}. We use a pose estimation model trained on COCO to predict 17 keypoints on ReID images, the predicted keypoints are shown in Figure~\ref{fig:pcb_pap_regions_c}. With these keypoints, we determine 9 regions \{\textit{head, upper torso, lower torso, upper leg, lower leg, foot, upper body, lower body, full body}\} for feature pooling. We call this strategy Part Aligned Pooling (PAP). Regions R1 $\sim$ R6 form a counterpart to compare with PCB in terms of alignment, while regions R7 $\sim$ R9 compensate for cases where the keypoint model fails to detect some local parts.

When occlusion or imperfect detection occurs, some parts may be invisible, \eg the feet are missing in the 3rd image of Figure~\ref{fig:pcb_pap_regions_b}. In this case, we do not pool feature for this part but use a $C$ dimensional zero vector $\overrightarrow{\mathbf{0}}$ as a replacement (before feeding it to the embedding layer), and loss from this part is ignored. The visibility aware loss is represented by
	\begin{equation}
		L^\text{id} = \sum_{p=1}^{P} L_p^\text{id} \cdot v_p,
	\end{equation} 
	where $v_p \in \{0, 1\}$ denotes the visibility of the $p$-th part of the image. During testing, we calculate image distance in an occlusion aware manner. For query image $I_q$, if the $i$-th part is invisible, the feature distance for this part between $I_q$ and all gallery images are ignored. Specifically, distance of a $\langle \text{\textit{query}},\ \text{\textit{gallery}} \rangle$ image pair $\langle I_q,\ I_g \rangle,\ g \in \{1,...,N\}$, where $N$ denotes gallery size, is calculated as
	\begin{equation} \label{eq:q_g_dist}
		D = \frac{\sum_{p=1}^{P} \text{cos\_dist}(e_p^q,\ e_p^g) \cdot v_p^q}{\sum_{i=1}^{P} v_i^q}
	\end{equation}
	where $e_p^q,\ e_p^g$ denote the $p$-th part embedding of $I_q$ and $I_g$ respectively, $v_p^q \in \{0, 1\}$ denotes visibility of the $p$-th part of $I_q$, and \textit{cos\_dist} denotes cosine distance. Note that if the $j$-th part is invisible on $I_g$ but visible on $I_q$, we maintain the result of $e_j^g = f_j(\overrightarrow{\mathbf{0}})$ when calculating Equation~\ref{eq:q_g_dist}.
	
\subsection{Part Segmentation Constraint}

\begin{figure}
    \centering
    \begin{subfigure}{1\linewidth}
        \centering
        \includegraphics[width=1\linewidth]{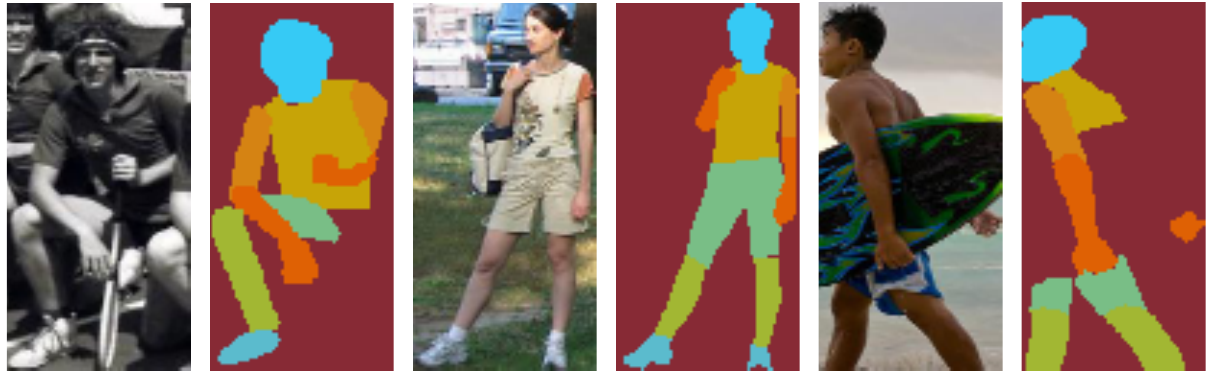}
        \caption{COCO Densepose Part Annotations (Fused into 7 Parts)}
        \label{fig:part_seg_label_a}
    \end{subfigure}%
    \\
    \begin{subfigure}{1\linewidth}
        \centering
        \includegraphics[width=1\linewidth]{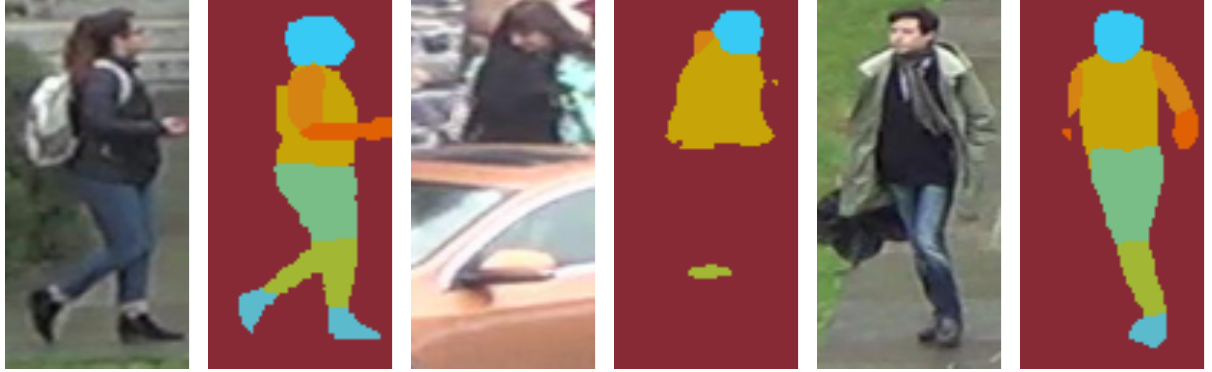}
        \caption{Pseudo Part Labels on ReID Datasets}
        \label{fig:part_seg_label_b}
    \end{subfigure}
    \caption{We fuse COCO Densepose~\cite{guler2018densepose} part labels into 8 classes, including \textit{background}, train a part segmentation model, and predict pseudo labels for ReID datasets.}
	\label{fig:part_seg_label}
\end{figure}

Since we perform fine-grained regional pooling on feature maps, we hope features pooled from different regions are distinct from each other, \ie with little redundancy. For example, we expect feature from R1 of Figure~\ref{fig:model} to be head centric, while feature from R6 is foot centric. However, as shown in Figure~\ref{fig:part_sim_b} and Figure~\ref{fig:part_sim_f}, there is still potential redundancy between part features. Motivated by this, we impose a part segmentation (PS) constraint on Conv5 feature maps, forcing a module to predict part labels from the feature maps. Our intuition is that, if the PS module is able to predict part labels from ReID feature maps, their localization ability (part awareness) are well maintained, and thus redundancy can be alleviated. Concretely, we connect a stride-2 $3 \times 3$ Deconv layer and then a $1 \times 1$ Conv layer to Conv5 feature maps, to predict part labels. The Deconv layer is for upsampling and $1 \times 1$ Conv layer is for pixel-wise classification. The supervision for part segmentation is pseudo labels predicted by a part segmentation model trained on COCO Densepose data. The COCO ground-truth part labels and ReID pseudo labels are illustrated in Figure~\ref{fig:part_seg_label}. 

The part segmentation loss on ReID feature is computed as follows, in a size-normalized manner.
	\begin{equation}
		L^\text{ps} = \frac{1}{K} \sum_{k=1}^{K} L^\text{ps}_k,
	\end{equation}
	where $K$ is number of part classes including \textit{background}, and $L^\text{ps}_k$ is the cross-entropy loss averaged inside the $k$-th part. The motivation of averaging inside each part before averaging across parts is to avoid large-size parts dominating the loss. It's important for small-size classes like \textit{foot} and \textit{head}, which also contain much discriminative information for ReID and should be equally attended.

\subsection{Multi-Task Training}

The source domain images are trained with both ReID loss and part segmentation loss. To adapt the model for target domain, we feed target-domain images to the network and train them with part segmentation loss. The total loss function is
	\begin{equation}
		L = L^\text{id}_S + \lambda_1 L^\text{ps}_S + \lambda_2 L^\text{ps}_T
	\end{equation}
	where $L^\text{id}_S$ denotes ReID loss of all source domain images, $L^\text{ps}_S$ PS loss of all source domain images, and $L^\text{ps}_T$ PS loss of all target domain images. $\lambda_1$ and $\lambda_2$ are loss weights for balancing different tasks, which are empirically set to 1.

\section{Experiments}

\subsection{Datasets and Evaluation Metrics}

We conduct our experiments on three large-scale person ReID datasets, Market1501~\cite{zheng2015scalable}, CUHK03~\cite{li2014deepreid} and DukeMTMC-reID~\cite{ristani2016performance,zheng2017unlabeled}. Two common evaluation metrics are used, Cumulative Match Characteristic (CMC)~\cite{gray2007evaluating} for which we report the Rank-1 accuracy, and mean Average Precision (mAP)~\cite{zheng2015scalable}. \textbf{Market1501} contains 12,936 training images from 751 persons, 29,171 test images from another 750 persons. \textbf{CUHK03} contains detected and hand-cropped images, both with 14,096 images from 1,467 identities. Following~\cite{zhong2017re}, we adopt the new train/test protocol with 767 training identities and 700 testing ones. We experiment on the detected images, which are close to real scenes. \textbf{DukeMTMC-reID} has the same format as Market1501, with 16,522 training images of 702 persons and 19,889  testing images of another 1110 persons.

\subsection{Implementation Details}

\textbf{Model.} We use ResNet-50~\cite{he2016deep} as the backbone, changing the stride of Conv5 from 2 to 1, as in PCB. Our GlobalPool baseline, Part Aligned Pooling, re-implemented PCB all use max pooling on Conv5. Embedding dimension is 512 for GlobalPool, and 256 for part based models. For fair comparison, we use cosine distance for GlobalPool baseline, average of part cosine distances for PCB, and Equation~\ref{eq:q_g_dist} for all PAP involved models. \textbf{Optimization.} We use SGD optimizer with a momentum of 0.9 and weight decay of 5e-4. Newly added layers has initial learning rate of 0.02, while layers to fine-tune use 0.01, all of which are multiplied by 0.1 after every 25 epochs. The training is terminated after 60 epochs. \textbf{Preprocessing.} Input images are resized to $w \times h = 128 \times 256$ for GlobalPool baseline, and $w \times h = 128 \times 384$ for part based models. Only random flipping is used as data augmentation for ReID training. Batch size is set to 32 and single-GPU training is used. When PS constraint is trained on cross-domain images, we resize them to $w \times h = 128 \times 384$ and also use batch size 32. The batch of ReID images and the batch of cross-domain images are fed to the network iteratively, but their gradients are combined before updating. \textbf{Pose Estimation.} The pose estimation model is trained on COCO, with AP 73.5\% on COCO 2017 val set when tested with ground truth bounding boxes. \textbf{COCO Part Labels.} The Densepose~\cite{guler2018densepose} dataset annotates 14 body parts, we fuse left/right parts and assign the \textit{hand} region to \textit{lower arm} class, getting 7 parts eventually. We use this transformed Densepose dataset to train a part segmentation model, adopting the recently proposed architecture DANet~\cite{fu2018dual}. With the trained part segmentation model, we predict part labels for ReID images.

\subsection{Effectiveness of PAP}

\begin{table*}
\footnotesize
\begin{center}
\begin{tabu} to 1\textwidth {l|X[c]|X[c]|X[c]|X[c]|X[c]|X[c]|X[c]|X[c]|X[c]}
\hline
& M$\to$M & C$\to$C & D$\to$D & M$\to$C & M$\to$D & C$\to$M & C$\to$D & D$\to$M & D$\to$C \\
\hline
\hline
GlobalPool & 	88.2 (71.3) &	42.4 (39.6) & 79.2 (61.9) & 10.7	 (9.3) & 38.7 (21.5) &	45.7 (21.8) &	32.5 (15.7) 	& 47.9 (21.6) &	9.1 (7.7) \\
PCB 		& 93.2 (81.1) &	65.2 (60.0) &	86.3 (72.7) &	8.9 (7.8)	& 42.9 (23.8)	& 52.1 (26.5) &	29.2 (15.2)	& 56.5 (27.7)	& 8.4 (6.9) \\
PAP-6P 	& 94.4 (84.2) & 68.1 (62.4) & 	85.6 (72.4) &	11.6 (9.9)	 & 47.6 (28.3)	& 54.6 (29.3)	& 33.9 (18.1) &	59.7 (31.4) &	9.2 (8.2) \\
PAP & 94.4 (84.5)	& 72.0 (66.2)	& 86.1 (73.3)	& 11.4 (9.9)	& 46.4 (27.9)	& 55.5 (30.0)	& 34.0 (17.9)	& 59.5 (30.6)	& 9.7 (8.0) \\
\hline
\end{tabu}
\end{center}
\caption{Effectiveness of PAP. Label \textbf{M}, \textbf{C} and \textbf{D} denote Market1501, CUHK03 and DukeMTMC-reID respectively. \textbf{M}$\to$\textbf{C} means training on M and testing on C, and so on. Score in each cell is Rank-1 (mAP) \%. \textbf{Following tables share this annotation.}}
\label{tab:comp_pcb_pap}
\end{table*}

We compare our model with PCB on three large scale datasets Market1501, CUHK03 and DukeMTMC-reID. The results are recorded in Table~\ref{tab:comp_pcb_pap}. We also list the performance of a vanilla baseline (GlobalPool), which performs global max pooling over the whole feature maps, obtaining one feature vector for each image. From the table we have the following analysis. (1) PCB surpasses GlobalPool by a large margin on source domain and most of the cross-domain testing cases. However, the performance drops when cross testing on CUHK03. (2) Cross-domain testing on CUHK03 is undesirably poor. (3) PAP-6P utilizes regions R1$\sim$R6 of Figure~\ref{fig:model} during both training and testing, which is a fair counterpart for PCB. We see that this aligned pooling consistently improves over PCB, especially 2.9\% Rank-1 for C$\to$C and 4.7\% for M$\to$D. (4) PAP utilizes all R1$\sim$R9 regions. It has competitive performance with PAP-6P on most of the cases, but increases 3.9\% Rank-1 for C$\to$C.

\subsection{Effectiveness of PS}

\begin{table*}
\footnotesize
\begin{center}
\begin{tabu} to 1\textwidth {l|X[c]|X[c]|X[c]|X[c]|X[c]|X[c]|X[c]|X[c]|X[c]}
\hline
& M$\to$M & C$\to$C & D$\to$D & M$\to$C & M$\to$D & C$\to$M & C$\to$D & D$\to$M & D$\to$C \\
\hline
\hline
PAP & 94.4 (84.5)	& 72.0 (66.2)	& 86.1 (73.3)	& 11.4 (9.9)	& 46.4 (27.9)	& 55.5 (30.0)	& 34.0 (17.9)	& 59.5 (30.6)	& 9.7 (8.0) \\
PAP-S-PS-SA & 94.5 (85.7) & 	71.4 (66.2) & 	86.9 (74.2) & 	13.6 (11.7) & 	50.2 (30.9) & 	58.4 (32.9) & 	38.4 (20.6) & 	60.6 (31.9) & 	11.1 (9.5) \\
PAP-S-PS & 94.6 (85.6)	 &	72.5 (66.8)	 &	87.5 (74.6)	 &	14.2 (12.8)	 &	51.4 (31.7)	 &	59.4 (33.3)	 &	39.3 (22.0) &		61.7 (32.9)	 &	11.4 (9.6) \\
PAP-ST-PS & - & - &	- &  21.4 (19.0) & 56.1 (36.0) & 66.4 (40.6) & 45.0 (26.4) & 66.1 (35.8) & 15.6 (13.8)  \\
\hline
\end{tabu}
\end{center}
\caption{Effectiveness of PS Constraint.}
\label{tab:comp_pap_ps}
\end{table*}

\begin{figure}
\begin{center}
   \includegraphics[width=1\linewidth]{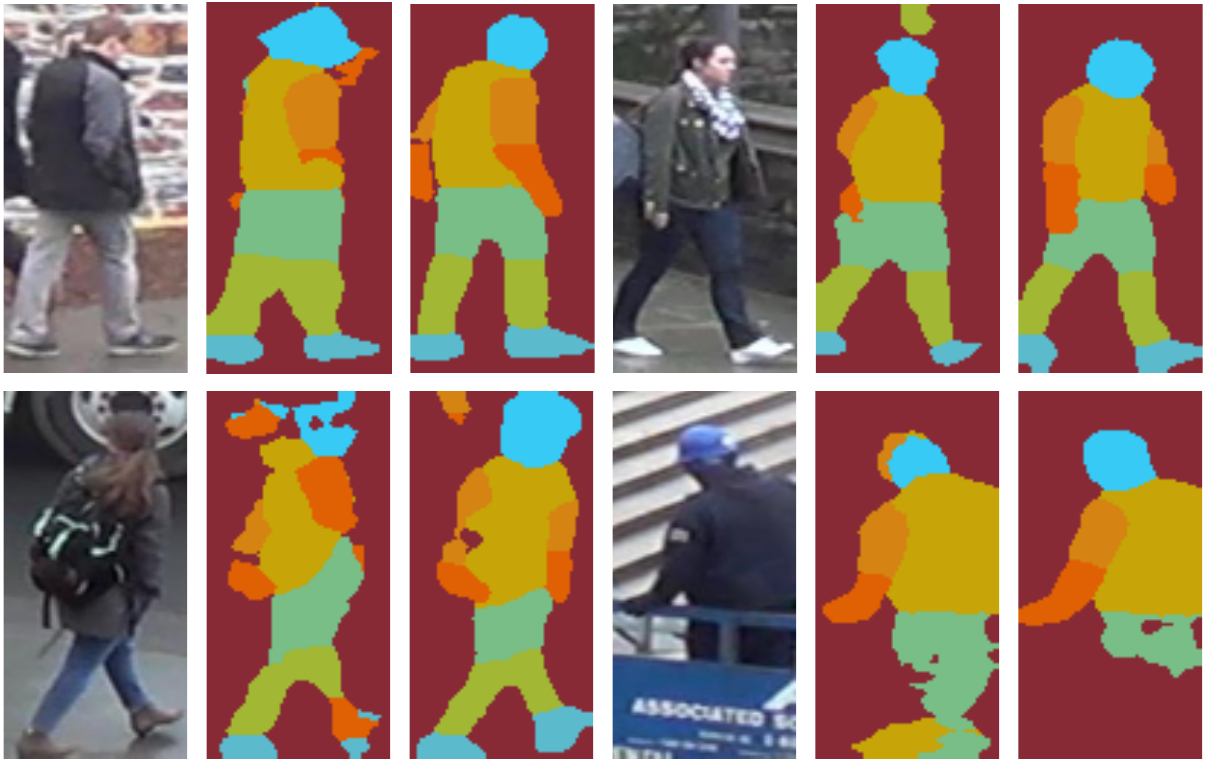}
\end{center}
   \caption{Part segmentation for target-domain images under C$\to$D setting. Each image is followed by part labels predicted from PAP-S-PS and PAP-ST-PS respectively. The PS constraint on target-domain images achieves alignment regularization.}
\label{fig:PAP_S_PS_vs_PAP_ST_PS_seg}
\end{figure}

We verify the effectiveness of our part segmentation constraint in Table~\ref{tab:comp_pap_ps}. \textbf{First}, we add PS to source domain images, denoted by PAP-S-PS. We see that it has obvious Rank-1 improvement for D$\to$D. Besides, it has larger superiority for cross-domain testing, improving Rank-1 by 2.8\%, 5.0\%, 3.9\%, 5.3\%, 2.2\%, 1.7\% for M$\to$C, M$\to$D, C$\to$M, C$\to$D, D$\to$M, D$\to$C, respectively. We can conclude that the PS constraint is effective under direct transfer setting. \textbf{Second}, we apply PS to both source and target domain images, which is denoted by PAP-ST-PS in the table. For example, for M$\to$C transfer, we train the model with loss $L^\text{id}_S$ on M, $L^\text{ps}_S$ on M, and $L^\text{ps}_T$ on C. The target domain PS constraint benefits pairwise transfer significantly, increasing Rank-1 by 7.2\%, 4.7\%, 7.0\%, 5.7\%, 4.4\%, 4.2\% for M$\to$C, M$\to$D, C$\to$M, C$\to$D, D$\to$M, D$\to$C, respectively. The part segmentation quality on target-domain images, with and without applying PS constraint to them, is also demonstrated in Figure~\ref{fig:PAP_S_PS_vs_PAP_ST_PS_seg}. \textbf{Besides}, we also verify the effect of balancing part sizes during calculating PS loss. We named the simple way of averaging over all pixel losses as PAP-S-PS-SA. Comparing PAP-S-PS-SA and PAP-S-PS, we can see consistent improvement by the balanced loss.

\subsection{Feature Similarity between Parts}

\begin{figure}
    \centering
    \begin{subfigure}{0.25\linewidth}
        \centering
        \includegraphics[width=1\linewidth]{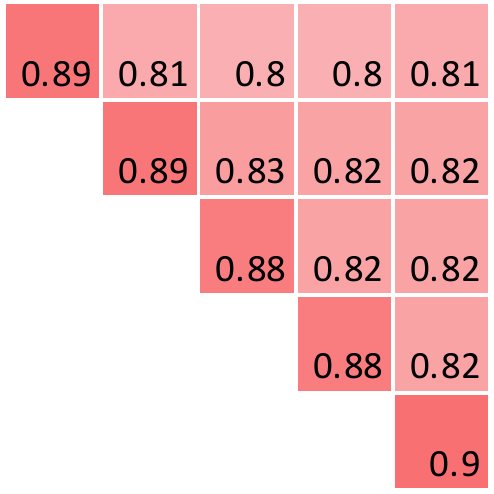}
        \caption{PCB}
        \label{fig:part_sim_a}
    \end{subfigure}%
    ~
    \begin{subfigure}{0.25\linewidth}
        \centering
        \includegraphics[width=1\linewidth]{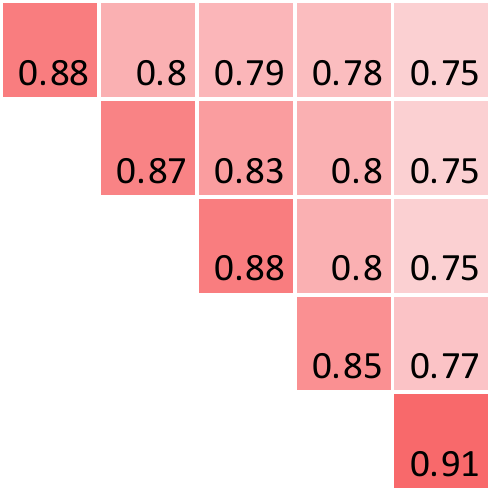}
        \caption{PAP}
        \label{fig:part_sim_b}
    \end{subfigure}%
    ~
    \begin{subfigure}{0.25\linewidth}
        \centering
        \includegraphics[width=1\linewidth]{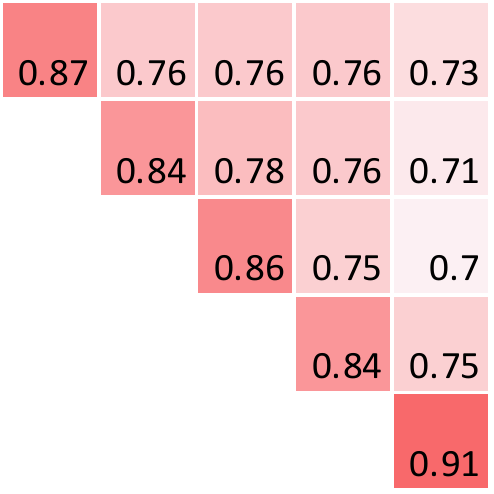}
        \caption{PAP-S-PS}
        \label{fig:part_sim_c}
    \end{subfigure}%
    ~
    \begin{subfigure}{0.25\linewidth}
        \centering
        \includegraphics[width=1\linewidth]{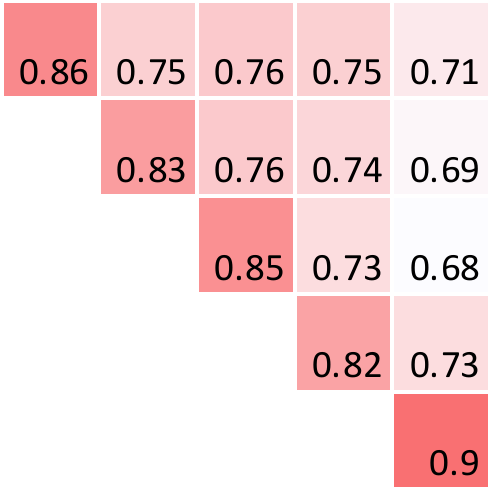}
        \caption{PAP-ST-PS}
        \label{fig:part_sim_d}
    \end{subfigure}%
    \\
    \begin{subfigure}{0.25\linewidth}
        \centering
        \includegraphics[width=1\linewidth]{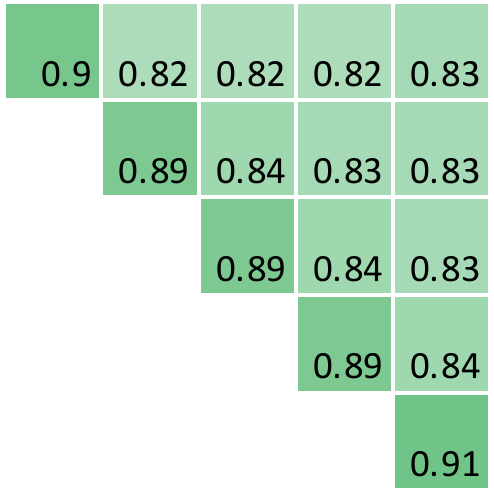}
        \caption{PCB}
        \label{fig:part_sim_e}
    \end{subfigure}%
    ~
    \begin{subfigure}{0.25\linewidth}
        \centering
        \includegraphics[width=1\linewidth]{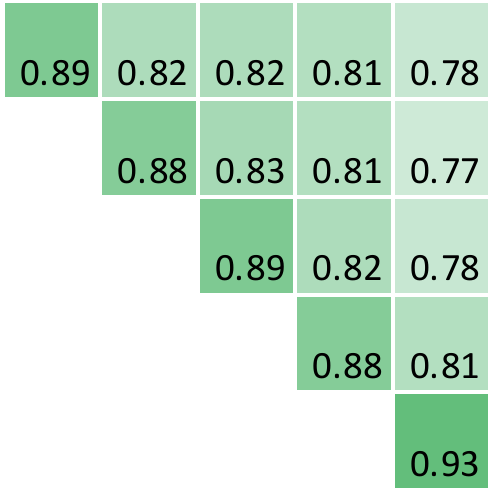}
        \caption{PAP}
        \label{fig:part_sim_f}
    \end{subfigure}%
    ~
    \begin{subfigure}{0.25\linewidth}
        \centering
        \includegraphics[width=1\linewidth]{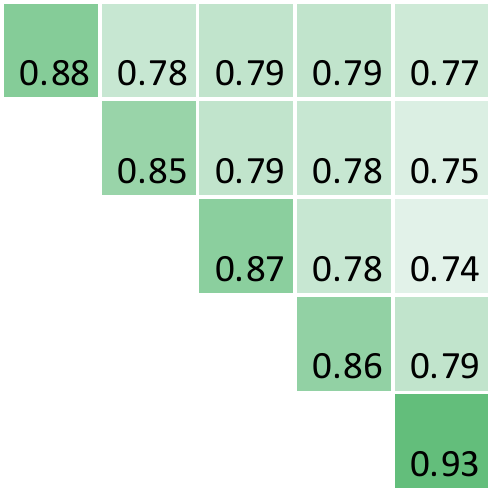}
        \caption{PAP-S-PS}
        \label{fig:part_sim_g}
    \end{subfigure}%
    ~
    \begin{subfigure}{0.25\linewidth}
        \centering
        \includegraphics[width=1\linewidth]{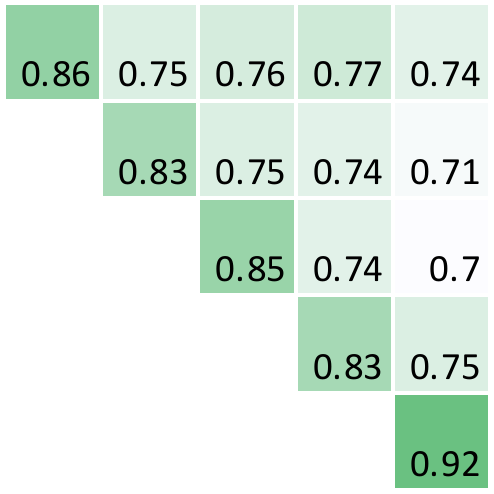}
        \caption{PAP-ST-PS}
        \label{fig:part_sim_h}
    \end{subfigure}%
    \caption{Cosine Similarity between Part Features $g_1\sim g_P$. (a) $\sim$ (d) are for C$\to$M and (e) $\sim$ (h) for C$\to$D setting. We only show the upper triangle (excluding diagonal line) of each symmetric matrix. For PAP, PAP-S-PS and PAP-ST-PS, we only show results between $g_1\sim g_6$.}
	\label{fig:part_sim}
\end{figure}

In this section, we analyze similarity between part features of a model. Although during testing, we use features ($e_1\sim e_P$) extracted from embedding layers, here we use those ($g_1 \sim g_P$) from Conv5 which is shared by all parts. We analyze features from PCB, PAP, PAP-S-PS and PAP-ST-PS. For each image, we calculate cosine similarity between its part features $g_1\sim g_P$, obtaining a $P\times P$ matrix. To analyze the statistical property, we average similarity matrices of images over the whole test set. The results of these models under C$\to$M and C$\to$D settings are shown in Figure~\ref{fig:part_sim}. Since the matrices are symmetric, we only show the upper triangles, excluding diagonal lines. Besides, for PAP, PAP-S-PS and PAP-ST-PS, we only show results between $g_1\sim g_6$. \textbf{First}, we notice that the results show large absolute values, with a minimum of 0.68. That's partially because the last step of ResNet Conv5 is a ReLU function which outputs non-negative values, which tends to increase the similarity value between vectors. \textbf{Second}, each part is mainly similar to its spatial neighbor(s). \textbf{Finally}, comparing Figure~\ref{fig:part_sim_a} $\sim$ \ref{fig:part_sim_d} or Figure~\ref{fig:part_sim_e} $\sim$ \ref{fig:part_sim_h}, we see that the proposed aligned pooling and part segmentation constraint reduce part similarity. We reckon that our model learns part centric features and reduces feature redundancy between parts.

\subsection{PS Constraint from COCO}

\begin{table*}
\footnotesize
\begin{center}
\begin{tabu} to 1\textwidth {l|X[c]|X[c]|X[c]|X[c]|X[c]|X[c]|X[c]|X[c]|X[c]}
\hline
& M$\to$M & C$\to$C & D$\to$D & M$\to$C & M$\to$D & C$\to$M & C$\to$D & D$\to$M & D$\to$C \\
\hline
\hline
PAP & 94.4 (84.5)	& 72.0 (66.2)	& 86.1 (73.3)	& 11.4 (9.9)	& 46.4 (27.9)	& 55.5 (30.0)	& 34.0 (17.9)	& 59.5 (30.6)	& 9.7 (8.0) \\
PAP-S-PS & 94.6 (85.6)	 & 72.5 (66.8)	 & 87.5 (74.6)	 & 14.2 (12.8)	 & 51.4 (31.7)	 & 59.4 (33.3) & 	39.3 (22.0)	 & 61.7 (32.9)	 & 11.4 (9.6) \\
PAP-C-PS & 92.9 (82.2)	& 64.9 (58.5)	& 84.9 (70.7)	& 18.3 (15.7)	& 54.3 (33.6)	& 66.1 (39.0)	& 44.6 (25.6)	& 64.1 (34.4)	& 12.7 (10.7)	 \\
PAP-StC-PS & 94.7 (84.9) & 70.1 (64.4) & 87.0 (73.4) & 19.1 (16.4)	 & 56.3 (35.1) & 65.5 (38.6) &	45.2 (26.1) & 65.2 (35.7) &	12.2 (10.5) \\
\hline
\end{tabu}
\end{center}
\caption{Promising Results of Introducing PS Constraint from COCO Dataset.}
\label{tab:ps_from_coco}
\end{table*}

In this section, we discuss the effect of training ReID with PS constraint from COCO images that come with ground-truth part labels. We train our PAP model along with the PS constraint computed on COCO images, which is denoted by PAP-C-PS in Table~\ref{tab:ps_from_coco}. It's interesting to see that, PAP-C-PS surpasses PAP-S-PS by a large margin in cross domain testing. However, the performance on source domain also drops obviously. To make the COCO images more consistent with ReID training set, we use a style transfer model SPGAN~\cite{deng2018image} to transfer COCO images to ReID style. In Table~\ref{tab:ps_from_coco}, PAP-StC-PS M$\to$C means training ReID on Market1501, as well as PS constraint on COCO images with Market1501 style, and testing on CUHK03. PAP-StC-PS C$\to$M \etc are similarly denoted. This style transformation reaches pleasing results, with source domain performance competitive to, and cross domain scores much higher than PAP-S-PS. Through this experiment, we show that with our PS constraint, it's feasible to achieve much better ReID model with the assistance of a publicly available part segmentation dataset. The gain of PAP-StC-PS over PAP-S-PS comes from either precise part labels or additional data regularization or both of them. To fairly compare pseudo vs. groud-truth part label, we require ReID datasets with part annotation, which we leave to future work.

\subsection{Complementarity to Existing Methods}

\begin{table}
\footnotesize
\begin{center}
\begin{tabu} to 1\linewidth {l|X[c]|X[c]}
\hline
& M$\to$D & D$\to$M \\
\hline
\hline
PCB 		& 42.9 (23.8) & 56.5 (27.7) \\
PCB-SPGAN & 48.0 (28.4) & 61.9 (31.1) \\
PAP-S-PS & 51.4 (31.7) & 61.7 (32.9) \\
PAP-S-PS-SPGAN & 56.2 (35.5) & 67.7 (37.3) \\
PAP-ST-PS & 56.1 (36.0) & 66.1 (35.8) \\
PAP-ST-PS-SPGAN & 61.5 (39.4) & 69.6 (39.3) \\
PAP-ST-PS-SPGAN-CFT & 67.7 (48.0) & 78.0 (51.6) \\
\hline
\end{tabu}
\end{center}
\caption{Complementarity to SPGAN~\cite{deng2018image} and Label Estimation~\cite{song2018unsupervised}.}
\label{tab:comp_to_existing_methods}
\end{table}

There have been some efforts on cross-domain ReID, including style transfer~\cite{deng2018image,wei2017person} and target-domain label estimation~\cite{liu2017stepwise,lv2018unsupervised,li2018unsupervised,song2018unsupervised}, \etc. In this section, we demonstrate the complementarity of our model to existing ones. \textbf{(1)} Style transfer methods typically translate source-domain images into target-domain style and then train a usual model on these generated images. We use the publicly available Market1501 and DukeMTMC-reID images generated by SPGAN~\cite{deng2018image} to conduct experiments. We train PCB, PAP-S-PS and PAP-ST-PS on these images and report the scores in Table~\ref{tab:comp_to_existing_methods}, denoted by PCB-SPGAN, PAP-S-PS-SPGAN and PAP-ST-PS-SPGAN, respectively. The direct cross-domain testing score of our PAP-S-PS model is already higher than PCB-SPGAN, which demonstrates the generalization advantage of our model. While comparing PAP-S-PS-SPGAN and PAP-ST-PS, we see that our target-domain PS constraint is competitive to GAN samples. Finally, integrating PAP-ST-PS and SPGAN brings further and large improvement, increasing Rank-1 by 5.4\% and 3.5\% for M$\to$D and D$\to$M respectively. This shows that seeing target-domain styles and our alignment (PS) constraint are pleasingly complementary to each other. \textbf{(2)} Following \cite{song2018unsupervised}, we use DBSCAN~\cite{ester1996density} to estimate identity labels and fine-tune our model in target domain. We utilize PAP-ST-PS-SPGAN for clustering and fine-tuning, which is denoted by PAP-ST-PS-SPGAN-CFT in Table~\ref{tab:comp_to_existing_methods}. We see that the label estimation method further improves the transfer scores significantly, increasing mAP by 8.6\%, 12.3\% for M$\to$D and D$\to$M respectively. We believe that the label estimation methods largely benefit from our model which already has good performance on target domain.

\subsection{Single-Domain Comparison with SOTA}

\begin{table*}
\footnotesize
\begin{center}
\begin{tabu} to 1\textwidth {l|c|c|c|c|c|c|c}
\hline
\multirow{2}{*}{Method} & \multirow{2}{*}{Publication} & \multicolumn{2}{c|}{Market1501} & \multicolumn{2}{c|}{CUHK03} & \multicolumn{2}{c}{DukeMTMC-reID} \\
\cline{3-8}
& & Rank-1    & mAP   & Rank-1 & mAP & Rank-1 & mAP \\
\hline
\hline
BoW+KISSME~\cite{zheng2015scalable} & ICCV15 & 44.4 & 20.8 & 6.4 & 6.4 & 25.1 & 12.2 \\
WARCA~\cite{jose2016scalable} & ECCV16 & 45.2 & - & - & - & - & - \\
\hline
SVDNet~\cite{sun2017svdnet} & ICCV17 & 82.3 & 62.1 & 41.5 & 37.3 & 76.7 & 56.8 \\
Triplet Loss~\cite{hermans2017defense} & arXiv17 & 84.9 & 69.1 & - & - & - & - \\
DaRe~\cite{wang2018resource} & CVPR18 &  86.4 & 69.3 & 55.1 & 51.3 & 75.2 & 57.4 \\
AOS~\cite{huang2018adversarially} & CVPR18 & 86.5 & 70.4 & 47.1 & 43.3 & 79.2 & 62.1 \\
DML~\cite{zhang2017deep} & CVPR18 & 87.7 & 68.8 & - & - & - & - \\
Cam-GAN~\cite{zhong2018camera} & CVPR18 & 88.1 &  68.7 & - & - & 75.3 & 53.5 \\
MLFN~\cite{chang2018multi} & CVPR18 & 90.0 & 74.3 & 52.8 & 47.8 & 81.0 & 62.8 \\
\hline
* PDC~\cite{su2017pose} & ICCV17 & 84.4 & 63.4 & - & - & - & - \\
* PSE~\cite{sarfraz2017pose} & CVPR18 &  87.7 & 69.0 & - & - & 79.8 & 62.0 \\
* PN-GAN~\cite{qian2017pose} & ECCV18 & 89.4 & 72.6 & - & - & 73.6 & 53.2 \\
* GLAD~\cite{wei2017glad} & MM17 & 89.9 & 73.9 & - & - & - & - \\
* PABR~\cite{suh2018part} & ECCV18 & 91.7 & 79.6 & - & - & 84.4 & 69.3 \\
* SPReID~\cite{kalayeh2018human} & CVPR18 & 92.5 & 81.3 & - & - &  84.4 & 71.0 \\
\hline
MSCAN~\cite{li2017learning} & CVPR17 & 80.3 & 57.5 & - & - & - & - \\
PAR~\cite{zhao2017deeply} & ICCV17 & 81.0 & 63.4 & - & - & - & - \\
JLML~\cite{li2017person} & IJCAI17 & 85.1 & 65.5 & - & - & - & - \\
HA-CNN~\cite{li2018harmonious} & CVPR18 & 91.2 & 75.7 & 41.7 & 38.6 & 80.5 & 63.8\\
AlignedReID~\cite{zhang2017alignedreid} & arXiv17 & 92.6 & \textcolor{blue}{82.3} & - & - & - & - \\
Mancs~\cite{wang2018mancs} & ECCV18 & 93.1 & \textcolor{blue}{82.3} & \textcolor{blue}{65.5} & \textcolor{blue}{60.5} & 84.9 & 71.8\\
\hline
PCB~\cite{sun2017beyond} & ECCV18 & 92.3 & 77.4 & 61.3 & 54.2 & 81.8 & 66.1 \\
PCB+RPP~\cite{sun2017beyond} & ECCV18 & \textcolor{blue}{93.8} & 81.6 & 63.7 & 57.5 & 83.3 & 69.2 \\
PCB (Our Implementation)	& - & 93.2 & 81.1 & 65.2 & 60.0 & \textcolor{blue}{86.3} & \textcolor{blue}{72.7} \\
\hline
* PAP (Ours) \dag & - & 94.4 & 84.5 & 72.0 & 66.2 & 86.1 & 73.3 \\
* PAP-S-PS (Ours) & - & \textcolor{red}{94.6} & \textcolor{red}{85.6} & \textcolor{red}{72.5} & \textcolor{red}{66.8} & \textcolor{red}{87.5} & \textcolor{red}{74.6} \\
\hline
\end{tabu}
\end{center}
\vspace{-1em}
\caption{Comparison with state-of-the-art methods on Market1501, CUHK03 (new protocol, \textit{detected} subset) and DukeMTMC-reID, under single-domain setting. Methods with * require keypoint and (or) segmentation assistance. In each column, the \textcolor{red}{1st} and \textcolor{blue}{2nd} highest scores (excluding methods with trailing \dag) are marked by \textcolor{red}{red} and \textcolor{blue}{blue}, respectively.}
\label{tab:single_domain_sota}
\end{table*}

Our comparison with state-of-the-art methods on Market1501, CUHK03 and DukeMTMC-reID, under single-domain setting, is listed in Table~\ref{tab:single_domain_sota}. Our PAP-S-PS model achieves state-of-the-art performance on three datasets, surpassing previous model PCB+RPP by 4.0\%, 9.3\%, 5.4\% mAP on Market1501, CUHK03 and DukeMTMC-reID, respectively. Compared with other keypoint or parsing assisted methods, our superiority is even more significant, which is attributed to our aligned pooling, each-part supervision and segmentation constraint.

\subsection{Cross-Domain Comparison with SOTA}

\begin{table*}
\footnotesize
\begin{center}
\begin{tabu} to 1\textwidth {l|c|c|c|c|c|c|c|c|c}
\hline
\multirow{2}{*}{Method} & \multirow{2}{*}{Publication} & \multicolumn{2}{c|}{M$\to$D} & \multicolumn{2}{c|}{D$\to$M} & \multicolumn{2}{c|}{C$\to$M} & \multicolumn{2}{c}{C$\to$D} \\
\cline{3-10}
& & Rank-1    & mAP   & Rank-1 & mAP & Rank-1 & mAP & Rank-1 & mAP \\
\hline
\hline
LOMO~\cite{liao2015person} & CVPR15 & 12.3 & 4.8 & 27.2 & 8.0 & - & - & - & - \\
BoW~\cite{zheng2015scalable} & ICCV15 & 17.1 & 8.3 & 35.8 &  14.8 & - & - & - & - \\
UMDL~\cite{peng2016unsupervised} & CVPR16 & 18.5 & 7.3 & 34.5 &  12.4 & - & - & - & - \\
CAMEL~\cite{yu2017cross} & ICCV17 & - & - & 54.5 & 26.3 & & & & \\
PUL~\cite{fan2018unsupervised} & TOMM18 & 30.0 & 16.4 & 45.5 & 20.5 & 41.9 & 18.0 & 23.0 & 12.0 \\
PTGAN~\cite{wei2017person} & CVPR18 & 27.4 & - & 38.6 & - & 31.5 & - & 17.6 & - \\
SPGAN~\cite{deng2018image} & CVPR18 & 41.1 &  22.3 & 51.5 & 22.8 & 42.3 &  19.0 & - & - \\
SPGAN+LMP~\cite{deng2018image} & CVPR18 & 46.4 & 26.2 & 57.7 & 26.7 & - & - & - & - \\

TJ-AIDL~\cite{wang2018transferable} & CVPR18 & 44.3 & 23.0 & 58.2 & 26.5 & - & - & - & - \\
HHL~\cite{zhong2018generalizing} & ECCV18 & \textcolor{blue}{46.9} & \textcolor{blue}{27.2} & \textcolor{blue}{62.2} & \textcolor{blue}{31.4} & \textcolor{blue}{56.8} & \textcolor{blue}{29.8} & \textcolor{blue}{42.7} & \textcolor{blue}{23.4} \\
\hline
GlobalPool (DT) \dag & - & 38.7 & 21.5 & 47.9 & 21.6 & 45.7 & 21.8 &	32.5 & 15.7 \\
PCB~\cite{sun2017beyond} (Our Imp., DT) \dag & ECCV18 & 42.9 & 23.8 & 56.5 & 27.7 & 52.1 & 26.5 &	 29.2 & 15.2  \\
\hline
PAP-S-PS (Ours, DT) \dag & - & 51.4 & 31.7 & 61.7 & 32.9 & 59.4 & 33.3 & 39.3 & 22.0 \\
PAP-ST-PS (Ours) & - & \textcolor{red}{56.1} & \textcolor{red}{36.0} & \textcolor{red}{66.1} & \textcolor{red}{35.8} & \textcolor{red}{66.4} & \textcolor{red}{40.6} & \textcolor{red}{45.0} & \textcolor{red}{26.4} \\
\hline
PAP-ST-PS-SPGAN (Ours) \dag & - & 61.5 & 39.4 & 69.6 & 39.3 & - & - & - & - \\
PAP-ST-PS-SPGAN-CFT (Ours) \dag & - & 67.7 & 48.0 & 78.0 & 51.6 & - & - & - & - \\
\hline
\end{tabu}
\end{center}
\vspace{-1em}
\caption{Comparison with state-of-the-art methods under cross-domain setting. In each column, the \textcolor{red}{1st} and \textcolor{blue}{2nd} highest scores (excluding methods with trailing \dag) are marked by \textcolor{red}{red} and \textcolor{blue}{blue}, respectively. \textbf{DT} means Direct Transfer.}
\label{tab:cross_domain_sota}
\end{table*}

We compare our model with state of the art under cross-domain setting, as reported in Table~\ref{tab:cross_domain_sota}. The direct transfer scores of our PAP-S-PS model is already competitive with previous domain adaptation methods that utilize target-domain images. As for our unsupervised cross-domain adaptation method PAP-ST-PS, it surpasses previous methods by a large margin. We surpass HHL~\cite{zhong2018generalizing} by 8.8\%, 4.4\%, 10.8\%, 3.0\% in mAP for M$\to$D, D$\to$M, C$\to$D, C$\to$M settings, respectively.

\section{Conclusion}

This work mainly verified the important role of alignment for cross-domain person ReID. We addressed model generalization and domain adaptation in the same effort. The proposed Part Aligned Pooling and Part Segmentation constraint not only effectively improve the generalization and adaptation of ReID model, but are also complementary to existing methods. The model components were analyzed with extensive experiments, and we also demonstrated its state-of-the-art performance in the literature. Future work includes (1) replacing keypoint with part segmentation maps for more precise region pooling as well as eliminating the need for the pose estimation model, (2) designing segmentation and ReID multi-task network, getting rid of additional segmentation model, (3) applying the alignment mechanism to occlusion scenarios.

{\small
\bibliographystyle{ieee}
\bibliography{2343.bib}

\begin{thebibliography}{10}\itemsep=-1pt

\bibitem{cao2017realtime}
Z.~Cao, T.~Simon, S.-E. Wei, and Y.~Sheikh.
\newblock Realtime multi-person 2d pose estimation using part affinity fields.
\newblock In {\em CVPR}, 2017.

\bibitem{chang2018multi}
X.~Chang, T.~M. Hospedales, and T.~Xiang.
\newblock Multi-level factorisation net for person re-identification.
\newblock In {\em CVPR}, 2018.

\bibitem{chen2017beyond}
W.~Chen, X.~Chen, J.~Zhang, and K.~Huang.
\newblock Beyond triplet loss: A deep quadruplet network for person
  re-identification.
\newblock In {\em CVPR}, 2017.

\bibitem{deng2018image}
W.~Deng, L.~Zheng, G.~Kang, Y.~Yang, Q.~Ye, and J.~Jiao.
\newblock Image-image domain adaptation with preserved self-similarity and
  domain-dissimilarity for person reidentification.
\newblock In {\em CVPR}, 2018.

\bibitem{ester1996density}
M.~Ester, H.-P. Kriegel, J.~Sander, X.~Xu, et~al.
\newblock A density-based algorithm for discovering clusters in large spatial
  databases with noise.
\newblock In {\em KDD}, 1996.

\bibitem{fan2018unsupervised}
H.~Fan, L.~Zheng, C.~Yan, and Y.~Yang.
\newblock Unsupervised person re-identification: Clustering and fine-tuning.
\newblock {\em TOMM}, 2018.

\bibitem{fu2018dual}
J.~Fu, J.~Liu, H.~Tian, Z.~Fang, and H.~Lu.
\newblock Dual attention network for scene segmentation.
\newblock {\em arXiv}, 2018.

\bibitem{gong2017look}
K.~Gong, X.~Liang, D.~Zhang, X.~Shen, and L.~Lin.
\newblock Look into person: Self-supervised structure-sensitive learning and a
  new benchmark for human parsing.
\newblock In {\em CVPR}, 2017.

\bibitem{gray2007evaluating}
D.~Gray, S.~Brennan, and H.~Tao.
\newblock Evaluating appearance models for recognition, reacquisition, and
  tracking.
\newblock In {\em PETS Workshop}, 2007.

\bibitem{guler2018densepose}
R.~A. G{\"u}ler, N.~Neverova, and I.~Kokkinos.
\newblock Densepose: Dense human pose estimation in the wild.
\newblock In {\em CVPR}, 2018.

\bibitem{he2016deep}
K.~He, X.~Zhang, S.~Ren, and J.~Sun.
\newblock Deep residual learning for image recognition.
\newblock In {\em CVPR}, 2016.

\bibitem{hermans2017defense}
A.~Hermans, L.~Beyer, and B.~Leibe.
\newblock In defense of the triplet loss for person re-identification.
\newblock {\em arXiv}, 2017.

\bibitem{huang2018adversarially}
H.~Huang, D.~Li, Z.~Zhang, X.~Chen, and K.~Huang.
\newblock Adversarially occluded samples for person re-identification.
\newblock In {\em CVPR}, 2018.

\bibitem{jose2016scalable}
C.~Jose and F.~Fleuret.
\newblock Scalable metric learning via weighted approximate rank component
  analysis.
\newblock In {\em ECCV}, 2016.

\bibitem{kalayeh2018human}
M.~M. Kalayeh, E.~Basaran, M.~G{\"o}kmen, M.~E. Kamasak, and M.~Shah.
\newblock Human semantic parsing for person re-identification.
\newblock In {\em CVPR}, 2018.

\bibitem{li2017learning}
D.~Li, X.~Chen, Z.~Zhang, and K.~Huang.
\newblock Learning deep context-aware features over body and latent parts for
  person re-identification.
\newblock In {\em CVPR}, 2017.

\bibitem{li2018unsupervised}
M.~Li, X.~Zhu, and S.~Gong.
\newblock Unsupervised person re-identification by deep learning tracklet
  association.
\newblock In {\em ECCV}, 2018.

\bibitem{li2014deepreid}
W.~Li, R.~Zhao, T.~Xiao, and X.~Wang.
\newblock Deepreid: Deep filter pairing neural network for person
  re-identification.
\newblock In {\em CVPR}, 2014.

\bibitem{li2017person}
W.~Li, X.~Zhu, and S.~Gong.
\newblock Person re-identification by deep joint learning of multi-loss
  classification.
\newblock In {\em IJCAI}, 2017.

\bibitem{li2018harmonious}
W.~Li, X.~Zhu, and S.~Gong.
\newblock Harmonious attention network for person re-identification.
\newblock In {\em CVPR}, 2018.

\bibitem{liao2015person}
S.~Liao, Y.~Hu, X.~Zhu, and S.~Z. Li.
\newblock Person re-identification by local maximal occurrence representation
  and metric learning.
\newblock In {\em CVPR}, 2015.

\bibitem{lin2014microsoft}
T.-Y. Lin, M.~Maire, S.~Belongie, J.~Hays, P.~Perona, D.~Ramanan,
  P.~Doll{\'a}r, and C.~L. Zitnick.
\newblock Microsoft coco: Common objects in context.
\newblock In {\em ECCV}, 2014.

\bibitem{liu2017stepwise}
Z.~Liu, D.~Wang, and H.~Lu.
\newblock Stepwise metric promotion for unsupervised video person
  re-identification.
\newblock In {\em ICCV}, 2017.

\bibitem{lv2018unsupervised}
J.~Lv, W.~Chen, Q.~Li, and C.~Yang.
\newblock Unsupervised cross-dataset person re-identification by transfer
  learning of spatial-temporal patterns.
\newblock In {\em CVPR}, 2018.

\bibitem{peng2016unsupervised}
P.~Peng, T.~Xiang, Y.~Wang, M.~Pontil, S.~Gong, T.~Huang, and Y.~Tian.
\newblock Unsupervised cross-dataset transfer learning for person
  re-identification.
\newblock In {\em CVPR}, 2016.

\bibitem{qian2017pose}
X.~Qian, Y.~Fu, W.~Wang, T.~Xiang, Y.~Wu, Y.-G. Jiang, and X.~Xue.
\newblock Pose-normalized image generation for person re-identification.
\newblock In {\em ECCV}, 2018.

\bibitem{ristani2016performance}
E.~Ristani, F.~Solera, R.~Zou, R.~Cucchiara, and C.~Tomasi.
\newblock Performance measures and a data set for multi-target, multi-camera
  tracking.
\newblock In {\em ECCV}, 2016.

\bibitem{ristani2018features}
E.~Ristani and C.~Tomasi.
\newblock Features for multi-target multi-camera tracking and
  re-identification.
\newblock In {\em CVPR}, 2018.

\bibitem{sarfraz2017pose}
M.~S. Sarfraz, A.~Schumann, A.~Eberle, and R.~Stiefelhagen.
\newblock A pose-sensitive embedding for person re-identification with expanded
  cross neighborhood re-ranking.
\newblock In {\em CVPR}, 2018.

\bibitem{Schroff_2015_CVPR}
F.~Schroff, D.~Kalenichenko, and J.~Philbin.
\newblock Facenet: A unified embedding for face recognition and clustering.
\newblock In {\em CVPR}, 2015.

\bibitem{song2018unsupervised}
L.~Song, C.~Wang, L.~Zhang, B.~Du, Q.~Zhang, C.~Huang, and X.~Wang.
\newblock Unsupervised domain adaptive re-identification: Theory and practice.
\newblock {\em arXiv}, 2018.

\bibitem{su2017pose}
C.~Su, J.~Li, S.~Zhang, J.~Xing, W.~Gao, and Q.~Tian.
\newblock Pose-driven deep convolutional model for person re-identification.
\newblock In {\em ICCV}, 2017.

\bibitem{suh2018part}
Y.~Suh, J.~Wang, S.~Tang, T.~Mei, and K.~M. Lee.
\newblock Part-aligned bilinear representations for person re-identification.
\newblock In {\em ECCV}, 2018.

\bibitem{sun2017svdnet}
Y.~Sun, L.~Zheng, W.~Deng, and S.~Wang.
\newblock Svdnet for pedestrian retrieval.
\newblock In {\em ICCV}, 2017.

\bibitem{sun2017beyond}
Y.~Sun, L.~Zheng, Y.~Yang, Q.~Tian, and S.~Wang.
\newblock Beyond part models: Person retrieval with refined part pooling.
\newblock In {\em ECCV}, 2018.

\bibitem{szegedy2015going}
C.~Szegedy, W.~Liu, Y.~Jia, P.~Sermanet, S.~Reed, D.~Anguelov, D.~Erhan,
  V.~Vanhoucke, and A.~Rabinovich.
\newblock Going deeper with convolutions.
\newblock In {\em CVPR}, 2015.

\bibitem{wang2018mancs}
C.~Wang, Q.~Zhang, C.~Huang, W.~Liu, and X.~Wang.
\newblock Mancs: A multi-task attentional network with curriculum sampling for
  person re-identification.
\newblock In {\em ECCV}, 2018.

\bibitem{wang2018transferable}
J.~Wang, X.~Zhu, S.~Gong, and W.~Li.
\newblock Transferable joint attribute-identity deep learning for unsupervised
  person re-identification.
\newblock In {\em CVPR}, 2018.

\bibitem{wang2018resource}
Y.~Wang, L.~Wang, Y.~You, X.~Zou, V.~Chen, S.~Li, G.~Huang, B.~Hariharan, and
  K.~Q. Weinberger.
\newblock Resource aware person re-identification across multiple resolutions.
\newblock In {\em CVPR}, 2018.

\bibitem{wei2017person}
L.~Wei, S.~Zhang, W.~Gao, and Q.~Tian.
\newblock Person transfer gan to bridge domain gap for person
  re-identification.
\newblock In {\em CVPR}, 2018.

\bibitem{wei2017glad}
L.~Wei, S.~Zhang, H.~Yao, W.~Gao, and Q.~Tian.
\newblock Glad: global-local-alignment descriptor for pedestrian retrieval.
\newblock In {\em ACM MM}, 2017.

\bibitem{xiao2018simple}
B.~Xiao, H.~Wu, and Y.~Wei.
\newblock Simple baselines for human pose estimation and tracking.
\newblock In {\em ECCV}, 2018.

\bibitem{xu2018attention}
J.~Xu, R.~Zhao, F.~Zhu, H.~Wang, and W.~Ouyang.
\newblock Attention-aware compositional network for person re-identification.
\newblock In {\em CVPR}, 2018.

\bibitem{yu2017cross}
H.-X. Yu, A.~Wu, and W.-S. Zheng.
\newblock Cross-view asymmetric metric learning for unsupervised person
  re-identification.
\newblock In {\em ICCV}, 2017.

\bibitem{yu2017divide}
R.~Yu, Z.~Zhou, S.~Bai, and X.~Bai.
\newblock Divide and fuse: A re-ranking approach for person re-identification.
\newblock In {\em BMVC}, 2017.

\bibitem{zhang2017alignedreid}
X.~Zhang, H.~Luo, X.~Fan, W.~Xiang, Y.~Sun, Q.~Xiao, W.~Jiang, C.~Zhang, and
  J.~Sun.
\newblock Alignedreid: Surpassing human-level performance in person
  re-identification.
\newblock {\em arXiv}, 2017.

\bibitem{zhang2017deep}
Y.~Zhang, T.~Xiang, T.~M. Hospedales, and H.~Lu.
\newblock Deep mutual learning.
\newblock In {\em CVPR}, 2018.

\bibitem{zhao2017deeply}
L.~Zhao, X.~Li, J.~Wang, and Y.~Zhuang.
\newblock Deeply-learned part-aligned representations for person
  re-identification.
\newblock In {\em ICCV}, 2017.

\bibitem{zheng2015scalable}
L.~Zheng, L.~Shen, L.~Tian, S.~Wang, J.~Wang, and Q.~Tian.
\newblock Scalable person re-identification: A benchmark.
\newblock In {\em ICCV}, 2015.

\bibitem{zheng2016person}
L.~Zheng, Y.~Yang, and A.~G. Hauptmann.
\newblock Person re-identification: Past, present and future.
\newblock {\em arXiv}, 2016.

\bibitem{zheng2017unlabeled}
Z.~Zheng, L.~Zheng, and Y.~Yang.
\newblock Unlabeled samples generated by gan improve the person
  re-identification baseline in vitro.
\newblock In {\em ICCV}, 2017.

\bibitem{zhong2017re}
Z.~Zhong, L.~Zheng, D.~Cao, and S.~Li.
\newblock Re-ranking person re-identification with k-reciprocal encoding.
\newblock In {\em CVPR}, 2017.

\bibitem{zhong2018generalizing}
Z.~Zhong, L.~Zheng, S.~Li, and Y.~Yang.
\newblock Generalizing a person retrieval model hetero-and homogeneously.
\newblock In {\em ECCV}, 2018.

\bibitem{zhong2018camera}
Z.~Zhong, L.~Zheng, Z.~Zheng, S.~Li, and Y.~Yang.
\newblock Camera style adaptation for person re-identification.
\newblock In {\em CVPR}, 2018.

\end{thebibliography}
}

\clearpage
\appendix
\section*{Supplementary Material}
\section{Training on MSMT17}

MSMT17 is a large and challenging ReID dataset proposed by Wei \etal~\cite{wei2017person}. A total of 126,441 bounding boxes of 4,101 identities are annotated, which involve 15 cameras, wide light varieties, and different weather conditions. To verify the effectiveness of our method, we train the models on MSMT17 with single-domain and cross-domain settings and report the results in Table~\ref{tab:train_on_msmt17}. (1) We see that PAP-6P improves PCB in Rank-1 accuracy by 4.1\%, 5.3\%, 1.7\%, 4.7\% for MS$\to$MS, MS$\to$M, MS$\to$C and MS$\to$D, respectively. (2) Comparing PAP and PAP-S-PS, we see that training with source domain PS constraint improves Rank-1 accuracy by 2.7\%, 3.4\%, 3.9\% for MS$\to$M, MS$\to$C and MS$\to$D transfer, respectively. (3) The comparison of PAP-S-PS and PAP-ST-PS shows the benefit of applying PS constraint to unlabeled target-domain images as domain adaptation. It increases Rank-1 accuracy of MS$\to$M, MS$\to$C and MS$\to$D transfer by 2.0\%, 3.3\% and 1.4\%, respectively. (4) Comparing PAP and PAP-C-PS, the PS constraint computed on COCO images increases Rank-1 accuracy by 5.5\%, 4.6\%, 3.7\% for MS$\to$M, MS$\to$C and MS$\to$D transfer, respectively. We eventually conclude that the enhanced alignment on source domain images achieves a more generalizable model, and that the alignment (PS) constraint on unlabeled target domain images serves as effective domain adaptation.

\begin{table*}
\small
\begin{center}
\begin{tabu} to 1\linewidth {l|X[c]|X[c]|X[c]|X[c]}
\hline
  & MS$\to$MS & MS$\to$M & MS$\to$C & MS$\to$D \\
\hline
GlobalPool & 69.4 (40.5) & 52.0 (25.7) & 13.4 (12.1) & 57.0 (36.1) \\
PCB & 74.0 (47.7) & 58.9 (30.6) & 14.3 (13.2) & 58.3 (38.2) \\
PAP-6P & 78.1 (51.2) & 64.2 (35.9) & 16.0 (14.9) & 63.0 (43.1) \\
PAP & 79.2 (52.9) & 63.7 (35.3) & 16.0 (15.2) & 63.5 (43.6) \\
PAP-S-PS & 80.8 (55.3) & 66.4 (37.9) & 19.4 (17.4) & 67.4 (46.4) \\
PAP-C-PS & 80.7 (53.9) & 69.2 (40.6) & 20.6 (18.6) & 67.2 (46.5) \\
PAP-ST-PS & - & 68.4 (40.4) & 22.7 (21.2) & 68.8 (49.6) \\
\hline
GoogleNet~\cite{szegedy2015going} & 47.6 (23.0) & - & - & - \\
PDC~\cite{su2017pose} & 58.0 (29.7) & - & - & - \\
GLAD~\cite{wei2017glad} & 61.4 (34.0) & - & - & - \\
\hline
\end{tabu}
\end{center}
\caption{Results of Models Trained on MSMT17. \textbf{MS}: MSMT17}
\label{tab:train_on_msmt17}
\end{table*}

\section{Influence of Embedding Size}

\begin{table*}[t]
\footnotesize
\begin{center}
\begin{tabu} to 1\textwidth {l|X[c]|X[c]|X[c]|X[c]|X[c]|X[c]|X[c]|X[c]|X[c]}
\hline
& M$\to$M & C$\to$C & D$\to$D & M$\to$C & M$\to$D & C$\to$M & C$\to$D & D$\to$M & D$\to$C \\
\hline
PAP-128 & 94.2 (84.1) & 68.6 (63.5) & 86.7 (73.3) & 12.0 (10.3) & 46.3 (27.4) & 56.2 (30.2) & 32.0 (17.5) & 59.6 (30.0) & 8.8 (7.8)  \\
PAP-256 & 94.4 (84.5)	& 72.0 (66.2)	& 86.1 (73.3)	& 11.4 (9.9)	& 46.4 (27.9)	& 55.5 (30.0)	& 34.0 (17.9)	& 59.5 (30.6)	& 9.7 (8.0) \\
PAP-384 & 94.5 (85.2) & 70.1 (65.0) & 86.0 (73.4) & 12.1 (10.8) & 46.4 (28.1) & 55.6 (30.4) & 36.1 (19.8) & 59.8 (30.8) & 8.8 (7.6)  \\
\hline
\end{tabu}
\end{center}
\caption{Influence of Embedding Size.}
\label{tab:comp_em_dim}
\end{table*}

We analyze our PAP model with embedding size (for each part) set to 128, 256 (used in the main paper) or 384. The results are reported in Table~\ref{tab:comp_em_dim}, denoted by PAP-128, PAP-256, PAP-384 respectively. We observe that these dimension sizes only have prominent difference for C$\to$C and C$\to$D, while staying competitive for other settings. 

\section{Details of Part Segmentation Model}

We use DANet~\cite{fu2018dual} to train a part segmentation model on COCO Densepose~\cite{guler2018densepose} data. \textbf{Model.} The backbone is ResNet-50. For simplicity, we do not use Channel Attention Module, thus there is only one loss term to optimize. The multi-dilation parameter is set to (2, 4, 8). \textbf{Dataset.} The COCO Densepose dataset contains 46,507 person bounding boxes for training and 2243 for validation. It annotates segmentation labels for 14 parts, \ie \{\textit{torso, right hand, left hand, left foot, right foot, right upper leg, left upper leg, right lower leg, left lower leg, left upper arm, right upper arm, left lower arm, right lower arm, head}\}. To make the segmentation model easier to train, we fuse left/right parts into one class and fuse \textit{hand} into \textit{lower arm}, getting 7 parts eventually. \textbf{Style Augmentation.} In our experiments, we find that model trained on COCO images has pleasing performance on COCO val set, but fails in some cases of ReID data, sometimes having noisy prediction. We hypothesize that low resolution of ReID images is a key factor. We try to blur COCO images, but the results do not improve obviously. To train a model most suitable for ReID datasets, we transform COCO images to Market1501, CUHK03 and DukeMTMC-reID styles respectively, using SPGAN~\cite{deng2018image}. We then train a segmentation model with the combination of original, Market1501-style, CUHK03-style, and DukeMTMC-reID-style COCO images, with 186,028 training images in total. We find this method obviously improves prediction on ReID datasets. Note that we use this same segmentation model to predict pseudo labels for MSMT17 images, without transferring COCO images to MSMT17 style. \textbf{Common Augmentation.} The original DANet model targets scene segmentation which tends to require high-resolution images, while we tackle person part segmentation with a bounding box input each time. So we can use much smaller images. We denote a variable \textit{base size} by $S_{base}=192$. For each image in the current batch, we randomly select a value in interval $[0.75 \times S_{base}, 1.25 \times S_{base}]$ as the shortest size and resize the image, without changing the $height / width$ ratio. Afterwards, the image is rotated by a random degree in range $[-10, 10]$. Denoting another variable \textit{crop size} by $S_{crop}=256$, if any image side is smaller than $S_{crop}$, we have to pad the image with zeros. After padding, we randomly crop out a $S_{crop} \times S_{crop}$ square region, which is normalized by ImageNet image mean and std before being fed to the network. Random horizontal flipping is also used for augmentation. \textbf{Optimization.} We use SGD optimizer, with learning rate 0.003, which is multiplied by 0.6 after every epoch. The training takes 5 epochs. The batch size is set to 16, and two GPUs are used for training. \textbf{Testing.} During testing, we simply resize each image to have shortest size as $S_{base}$, \ie 192, while keeping the aspect ratio. No cropping or any other augmentation is applied. The final pixel accuracy on COCO val set (original COCO images, without changing style) is 0.903, and mIoU is 0.668. \textbf{Future Work.} We would like to utilize common data augmentation during testing, like flipping, cropping, and multi-scale testing, to achieve segmentation labels of higher quality.

\section{Details of Label Estimation}

The common practice of label estimation includes (1) training a ReID model on source domain, (2) extracting feature on target-domain unlabeled images and computing sample distance, (3) estimating pseudo labels using clustering methods, (4) fine-tuning the ReID model on these pseudo labels. Step (2) $\sim$ (4) can be repeated for times. Following Song \etal~\cite{song2018unsupervised}, we use DBSCAN~\cite{ester1996density} clustering method. The distance metric used is the cosine distance computed by our model, \ie Equation 3 of the main paper. Following \cite{ester1996density}, we set the clustering threshold $\epsilon$ to the statistical value computed from the distance matrix. For simplicity, we do not update our distance matrix by means of re-ranking or weighted fusion. Besides, the only data augmentation used in fine-tuning is flipping, without random erasing, \etc. During fine-tuning, we keep the model structure the same as training on source domain, except the new classifiers. The fine-tuning learning rate is set to 0.001 and 0.002 for original and classifier parameters, respectively. We find it sufficient to only fine-tune the model for 5 epochs without repetition. Although being important for ReID, exploring the best practice of clustering and fine-tuning is outside the scope of this paper.

\end{document}